\documentclass{article}

\PassOptionsToPackage{numbers, compress}{natbib}
\usepackage[preprint]{neurips_2021}

\usepackage[utf8]{inputenc} 
\usepackage[T1]{fontenc}    
\usepackage{hyperref}       
\usepackage{url}            
\usepackage{booktabs}       
\usepackage{amsfonts}       
\usepackage{nicefrac}       
\usepackage{microtype}      
\usepackage{xcolor}         
\usepackage{graphicx}
\usepackage{subcaption}

\title{Discriminator Synthesis: On reusing the other half of Generative Adversarial Networks}

\author{%
  Diego Porres\\
Computer Vision Center (CVC), Universitat Aut\`onoma de Barcelona\\
  Barcelona, Spain, 08193 \\
  \texttt{dporres@cvc.uab.es} \\
}

\begin{document}

\maketitle

\begin{abstract}
	Generative Adversarial Networks have long since revolutionized the world of computer vision and, tied to it, the world of art. Arduous efforts have gone into fully utilizing and stabilizing training so that outputs of the Generator network have the highest possible fidelity, but little has gone into using the Discriminator after training is complete. In this work, we propose to use the latter and show a way to use the features it has learned from the training dataset to both alter an image and generate one from scratch. We name this method Discriminator Dreaming, and the full code can be found at \url{https://github.com/PDillis/stylegan3-fun}.
\end{abstract}

\section{Introduction}

Generative adversarial networks (GAN) \cite{GAN} have, irrevocably, altered the course of machine-aided art. Some architectures are better suited for the task \cite{dcgan, biggan, cyclegan}, mainly due to the eventual ease of use and evaluation of these models, as well as the availability of data and compute. However, the main focus in the literature has been to increase the fidelity and stabilization of the images being generated \cite{progan,SGAN,SGAN2,SGAN2-ADA} and, while the Discriminator has to also increase its capability in order for these larger networks to be trained, it is usually left unexplored if not outright discarded.

While previous efforts have gone into using the Discriminator as part of an optimization process, either to use it in order to generate images it deems to be fake \cite{terence_uncanny} or to use it alongside another classifier to update the weights of the Generator \cite{terence_transformingGANs}, more effort can be done here, especially to use these large networks in an artistic manner. These networks have learned unique features from both the real and fake dataset, and they are simply being thrown away.

This work has two main sources of inspiration. The first is by Robbie Barrat who used the Discriminator from a trained GAN to optimize the tile placements of a nude portrait \cite{robbie_barrat}. The second is the recent global effort to use large pre-trained models such as CLIP \cite{CLIP}, which would not have been able to be trained by an individual researcher or artist. Two notable examples are the Big Sleep \cite{advadnoun} and Control The Soul \cite{rivers}, who use CLIP alongside BigGAN and StyleGAN, respectively, to generate images from prompts. 

Even though the features learned by a Discriminator may not be used for other downstream classification tasks (e.g., such as the models trained on ImageNet \cite{imagenet} which are then fine-tuned for other tasks), it is still worth exploring the learned features, especially for artistic endeavors. Indeed, what warrants as a 'fake' or 'real' image for the network itself will not translate to logical human sight features per se, but nevertheless these may still hold beauty.

We can use the Discriminator in a variety of ways. For example, we could use it in lieu of a VGG network \cite{vgg} for style transfer \cite{styletransfer}, or when projecting an image using the Generator of the same network (or even using another Generator altogether). Staying within StyleGAN, we could use it as a sort of feedback loop with the Generator, as the shape of the fully-connected layer before the final output (that is, the output shape of \texttt{D.b4.fc}) has the same shape as the latent space $\mathcal{Z}$. 

As a proof of concept, we use StyleGAN2's Discriminator instead of the Inception network \cite{inception} for Inceptionism/DeepDream \cite{deepdream}. We provide a starter code on how to extract the intermediate features of the Discriminator (with residual \cite{resnet} connections) of StyleGAN2, in particular, its ADA (PyTorch) version, but is also compatible with StyleGAN3's \cite{sgan3} repository and models. Our code is built upon the DeepDream-PyTorch repository \cite{deepdreampytorch} and we name our algorithm Discriminator Dreaming.

\section{Discriminator Dreaming}\label{discdream}

The following are some preliminary results we have obtained, though the final code is subject to change. For this work, we used a vast collection of available pre-trained models, some trained by third parties or provided by the authors of the original papers, others trained by ourselves. These official models were introduced in StyleGAN2 and StyleGAN2-ADA papers, and can be found in their respective GitHub repositories. These include: at $1024$ resolution, FFHQ and MetFaces; at $512$ , FFHQ-512, Cars, and AFHQ-Cat; and at $256$, FFHQ-256, LSUN Horse, and LSUN Church.

Other models include, at $1024$ resolution, MinecraftGAN, a model trained with images taken from a first-person POV in the videogame Minecraft throughout different weathers and biomes \cite{minecraftgan}. At $512$ resolution, the Huipiles model was trained with images from huipiles from Guatemala and Mexico \cite{huipiles}. At $256$ resolution, Doors is a model trained on art nouveau doors found in Barcelona \cite{doorgan}.

\begin{figure}
	\centering
	\includegraphics[width=\textwidth]{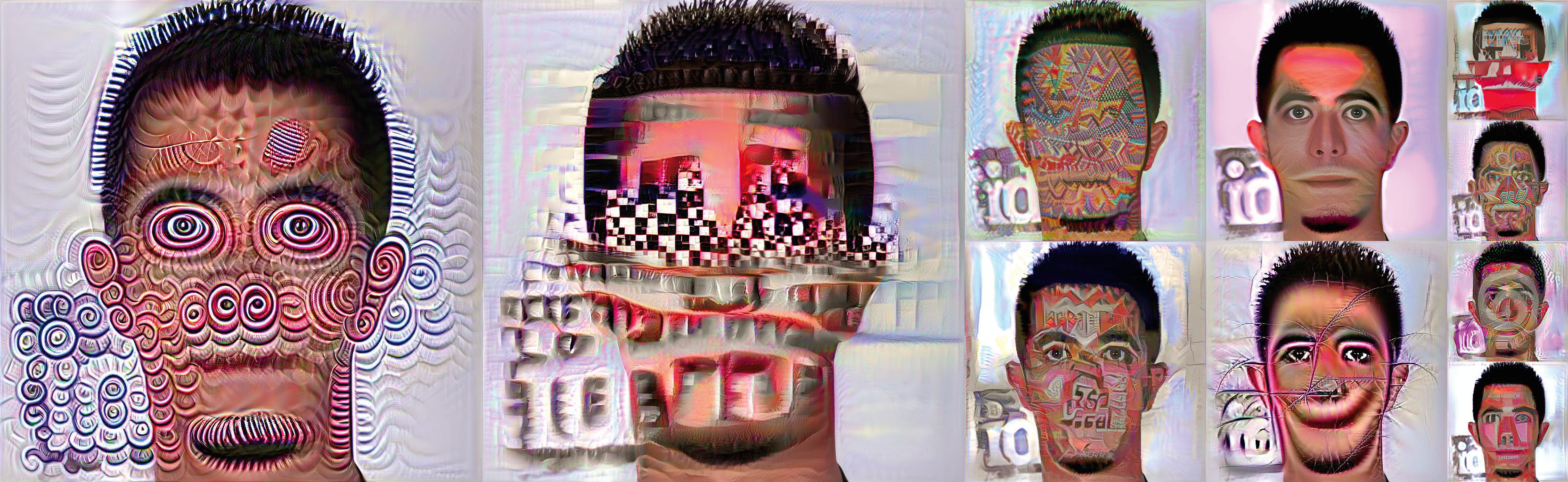}
	\caption{Discriminator Dreaming with StyleGAN2 models of different resolutions. From left to right and top to bottom: ($1024$ resolution) MetFaces and MinecraftGAN; ($512$ resolution) Huipiles, FFHQ-512, Cars, and AFHQ-Cat; ($256$ resolution) Doors, FFHQ-256, LSUN Horse, and LSUN Church. Results shown to scale. Best viewed in color.}
	\label{fig:main}
\end{figure}

In Figure \ref{fig:main}, we can see the result of Discriminator Dreaming using different layers for various models. We use, as a starting image, \href{https://drive.google.com/file/d/1vqQHRarSmzeXmhyYwrbmXQUTdZYmsICF/view?usp=sharing}{\texttt{00012.png}} from the FFHQ dataset introduced in StyleGAN. This is using a single frame, so for a video results, refer to Appendix \ref{video}. As per this section, for future work, we could set the zoom, rotation, and translation to be variable instead of fixed, by e.g. using an audio-reactive code \cite{audioreactive_sgan}. 

\subsection{Limitations}

Due to the iterative nature of this algorithm, it is slow to synthesize images, so perhaps more efficient algorithms could be used. Resizing the octaves (see Appendix \ref{appendix}) usually makes the code run twice as slow on a RTX 2070, making it less energy-efficient as a trained Generator. However, if we do not resize the octaves, then the image loses its sharpness, resulting in fuzzy images, as well as sometimes giving us vastly different results. Thus, a lot of tweaking and exploration will be needed in order to get the desired result. See Figures \ref{fig:1024_noresize} and \ref{fig:1024_resize} for a difference when resizing or not the octaves.

\begin{ack}
Diego Porres acknowledges the financial support to perform his Ph.D. given by the grant PRE2018-083417. 

\end{ack}

\section*{Ethical implications}

This work is meant for purely artistic use. We have limited our work to using StyleGAN2 and StyleGAN3's Discriminator, for which usually the available pre-trained models are based on publicly-available datasets. However, the trained Discriminator may have learned the distribution of images it was trained on for which we might not have the rights to. Fully utilizing the Machine Learning models we train is what this work strives towards, as the environmental implication of training each from scratch and then discarding half of the network should not be so easily dismissed.

\bibliography{references}


\appendix

\section{Appendix}\label{appendix}

\paragraph{Discriminator architecture}

A full breakdown of StyleGAN2's Discriminator architecture is beyond the scope of this work, though we refer the interested reader to its respective paper. In short, we will be mostly interested in the two main convolutional layers per block (e.g., \texttt{D.b1024.conv0} and \texttt{D.b1024.conv1} for the block at $1024\times1024$ scale). There will be $\log_{2}{(\texttt{D.img\_resolution}/4)}$ blocks in total, and the last one at $4\times4$ scale will have a minibatch standard-deviation layer, a convolutional layer, and two fully-connected layers, the latter one giving the actual output of the Discriminator (i.e., the same as if passing \texttt{D(input\_image, None)} to an unconditional model).

\begin{figure}
	\centering
	\includegraphics[width=\textwidth]{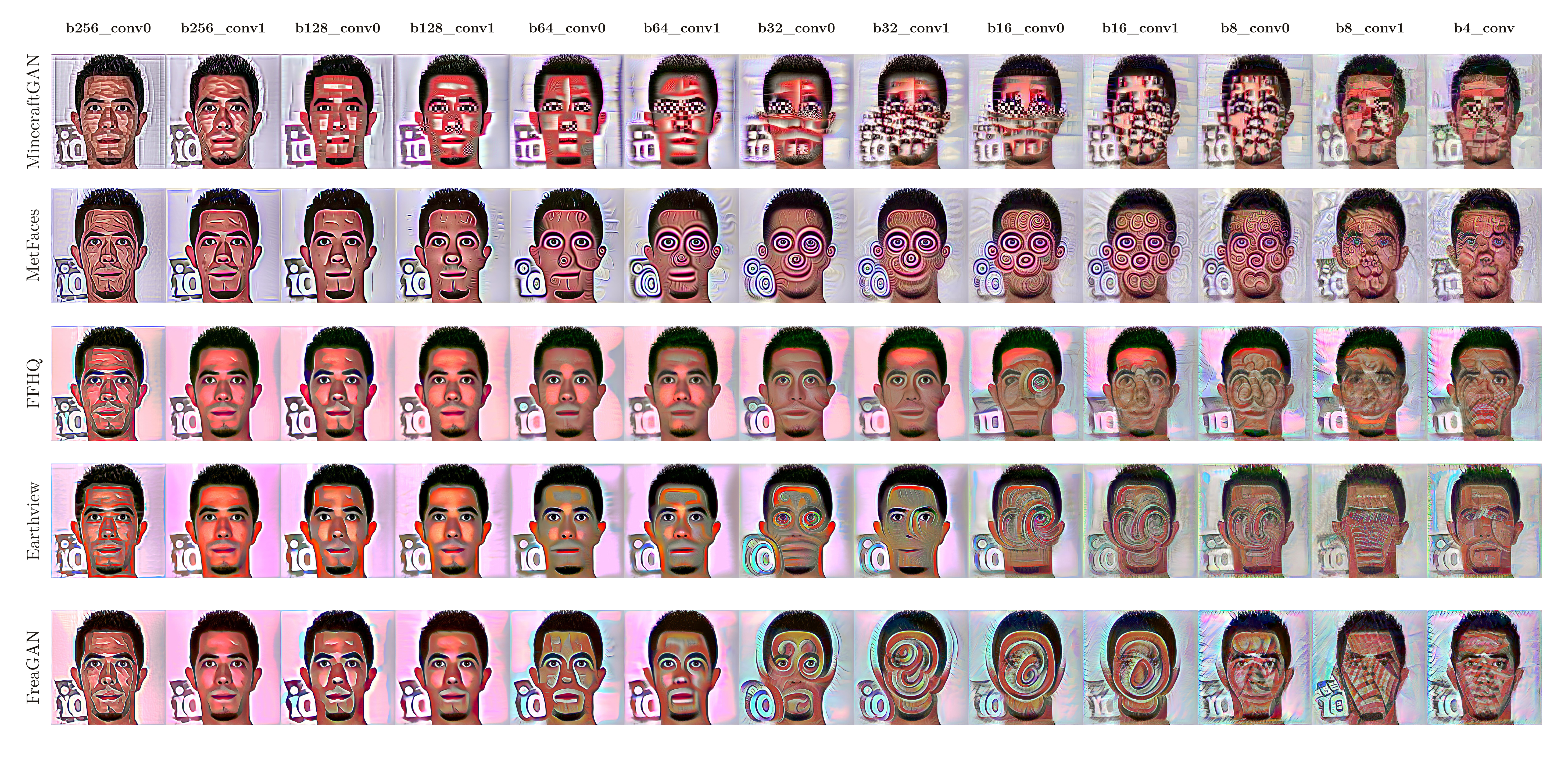}
	\caption{Discriminator Dreaming with different models of $1024\times1024$ resolution, without resizing the octaves. Columns denote the individual layer we are using to do the dreaming (starting from block $256\times256$ down to $4\times4$) and rows the models used. We use image \texttt{00012.png} as a starting image. Zoom in for details. Best viewed in color.}
	\label{fig:1024_noresize}
\end{figure}

\begin{figure}
	\centering
	\includegraphics[width=\textwidth]{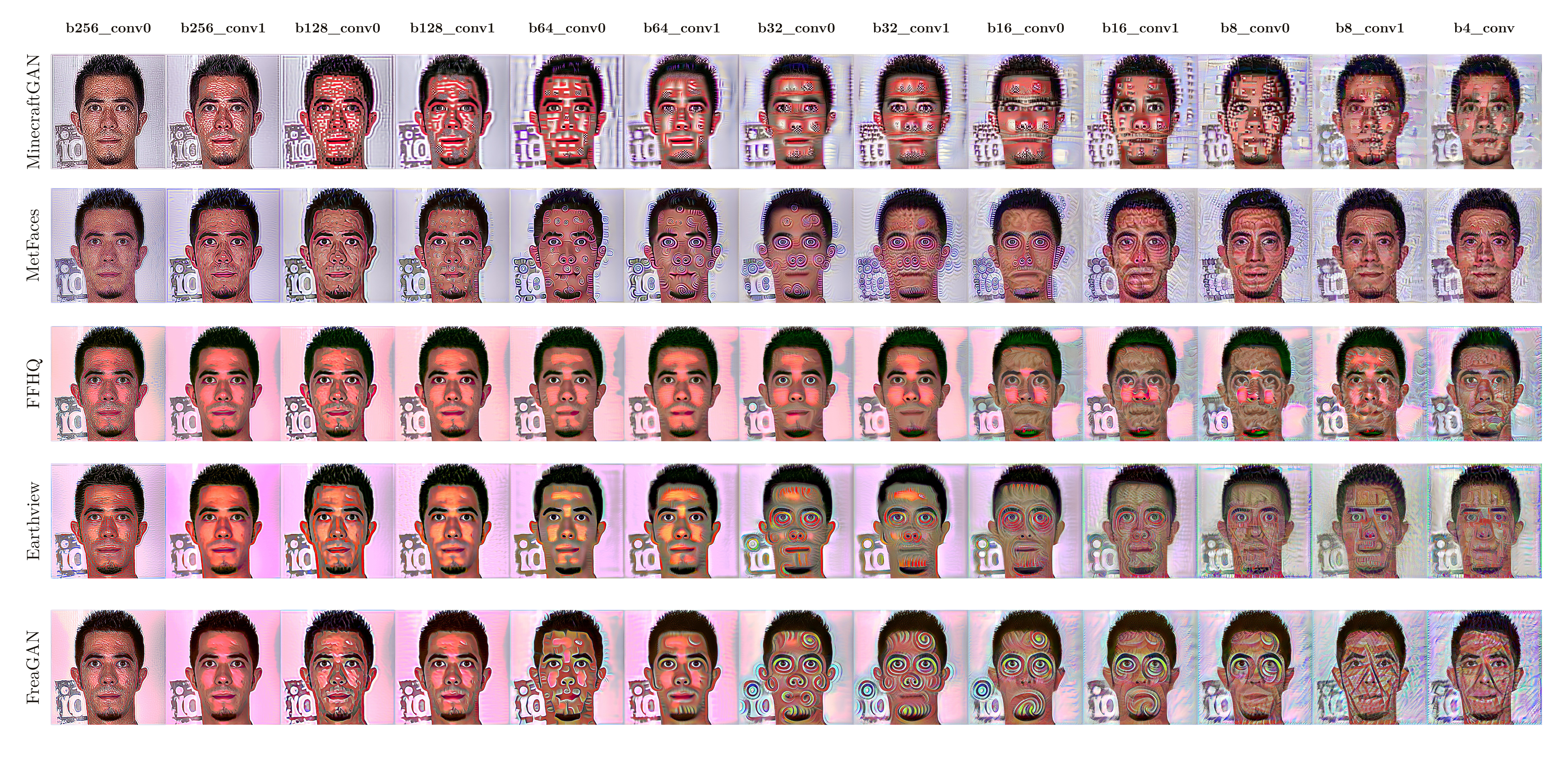}
	\caption{Discriminator Dreaming with different models of $1024\times1024$ resolution, plus resizing the octaves. Columns denote the individual layer we are using to do the dreaming (starting from block $256\times256$ down to $4\times4$) and rows the models used. We use image \texttt{00012.png} as a starting image. Zoom in for details. Best viewed in color.}
	\label{fig:1024_resize}
\end{figure}

\paragraph{Discriminator Dreaming with intermediate layers}
We show the preliminary results of Discriminator Dreaming with a StyleGAN2 Discriminator. In Figures \ref{fig:1024_noresize} and \ref{fig:1024_resize} we show results using the intermediate layers: from blocks at scales from $256\times256$ down to $4\times4$. In the selected models, the blocks at higher scale generally did not produce meaningful results, and the last fully-connected layers added little of interest, compared to the convolutional layer in the block at scale $4\times4$. However, all of the available layers can be accessed and used by the user if so is desired.

We show results with two additional models at $1024\times1024$ resolution. Earthview is a model trained with images from Earth View from Google, which are pre-selected satellite images of the most beautiful landscapes in Google Earth \cite{earthview}. FreaGAN is a model by Derrick Schultz which was trained on Frea Buckler's art \cite{dvs}. 

For both of these networks, we can see in Figures \ref{fig:1024_noresize} and \ref{fig:1024_resize} that they were trained via transfer learning from FFHQ, as the results when using the convolutional layers of the larger blocks are quite similar. This makes sense, as the training of these models was limited compared to that of FFHQ, mainly due to a difference in the size of each dataset, so the gradients were mostly concentrated in the lower resolution blocks.

\paragraph{Resizing octaves}

We can apply the algorithm to a single image, but that would rarely produce significant amount of alteration. In order to affect different layers of granularity of the image, we can apply Discriminator Dreaming at different scales by reducing the size of the image and adding this loss to the overall image. Each scale of the image is referred to as an octave.

However, given that our Discriminator has pretty aggressive downscaling per block (each block lowers the resolution by a factor of 2), we might eventually encounter errors of size mismatch. To solve this, we can simply resize each octave back to the original resolution. This will make the code run slower, but we also obtain different results, as can be seen in Figures \ref{fig:1024_noresize} and \ref{fig:1024_resize} (using the same parameters, except whether or not we resize the octaves).

\paragraph{Random image} 

\begin{figure}[htp]
	\begin{subfigure}{0.3\textwidth}
		\includegraphics[width=\linewidth]{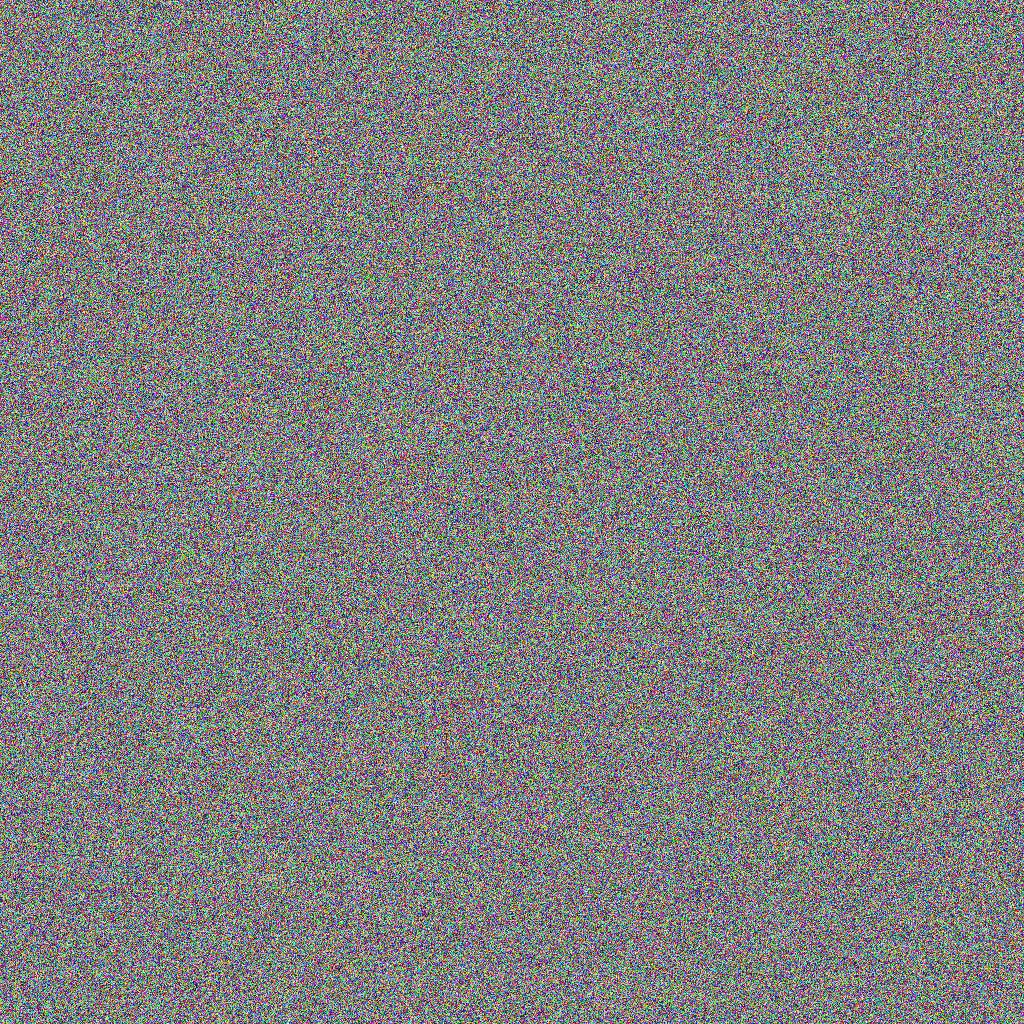}
		\caption{Random noise image (seed 0)}
	\end{subfigure}
	\hspace*{\fill}
	\begin{subfigure}{0.3\textwidth}
		\includegraphics[width=\linewidth]{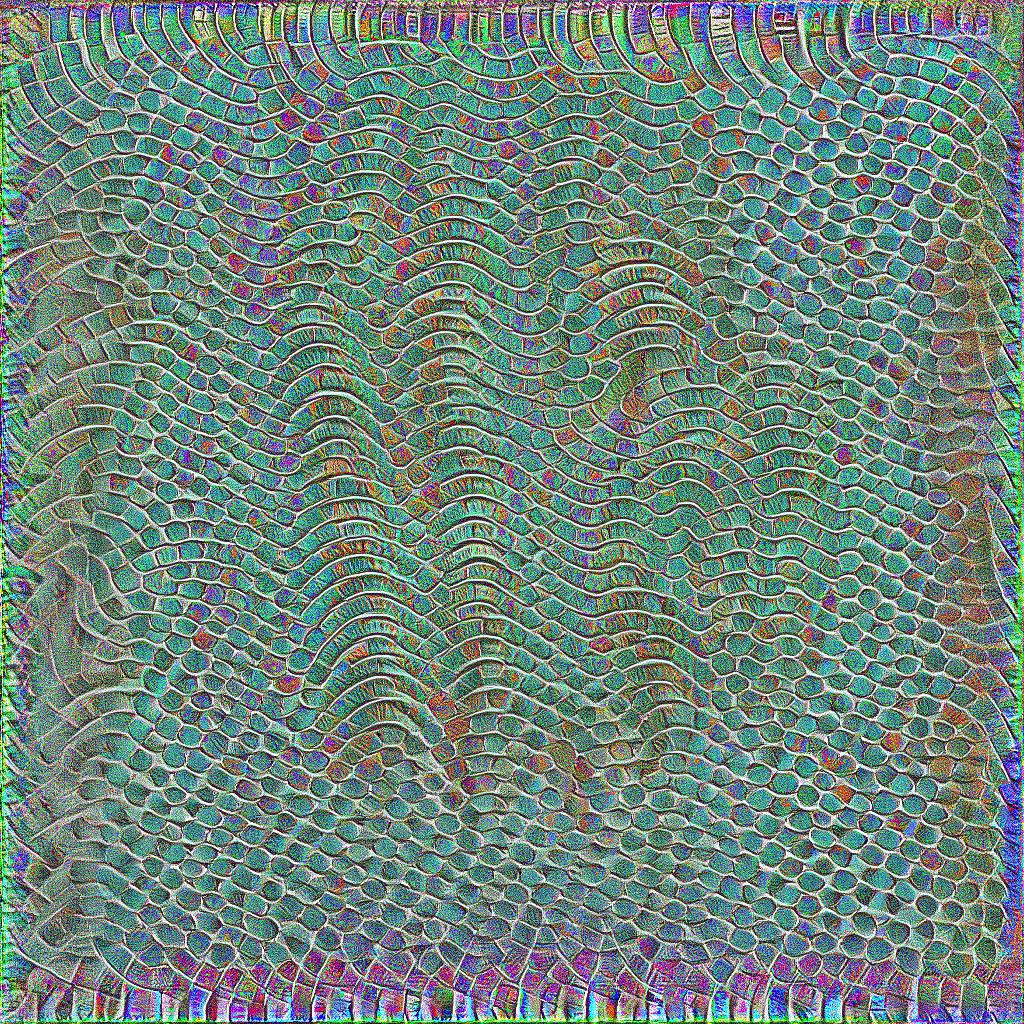}
		\caption{\texttt{b16\_conv1}}
	\end{subfigure}
	\hspace*{\fill}
	\begin{subfigure}{0.3\textwidth}
		\includegraphics[width=\linewidth]{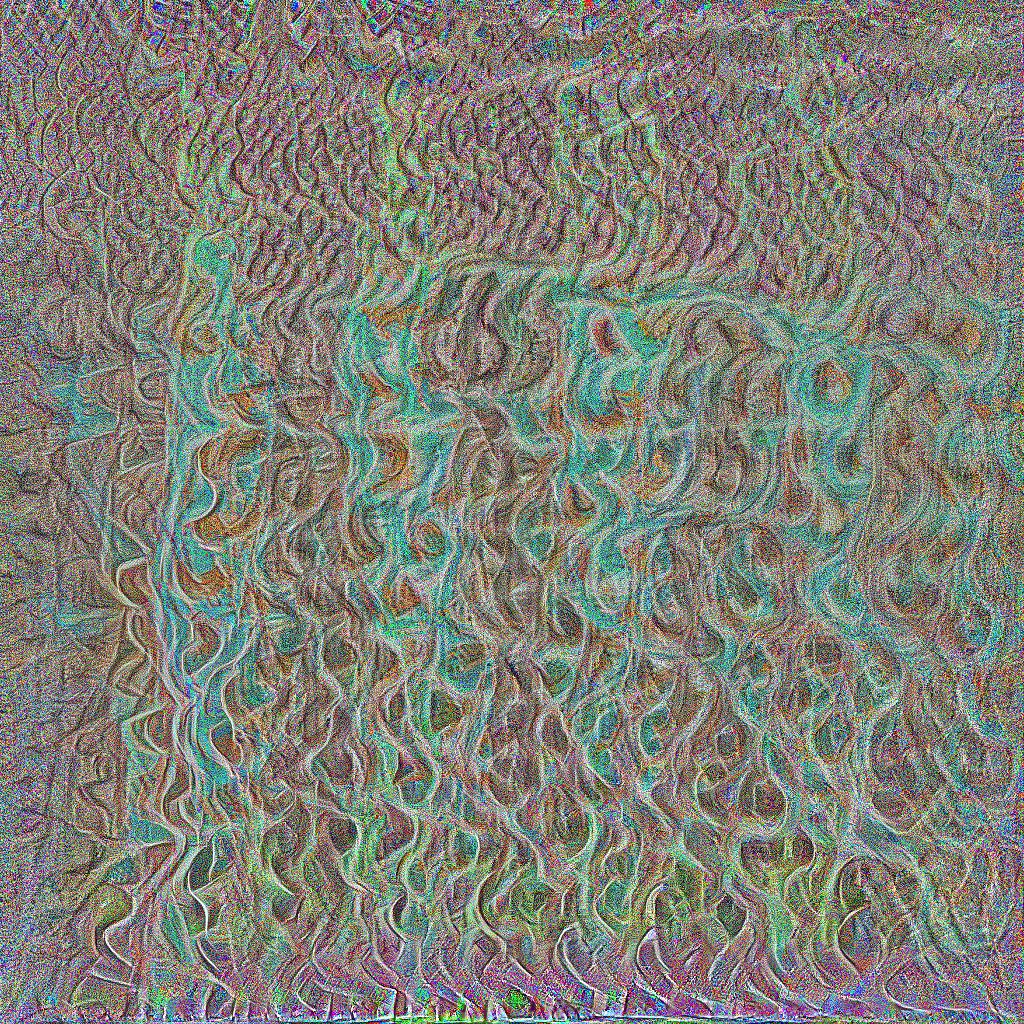}
		\caption{\texttt{b8\_conv1}}
	\end{subfigure}
	
	\bigskip
	
	\begin{subfigure}{0.3\textwidth}
		\includegraphics[width=\linewidth]{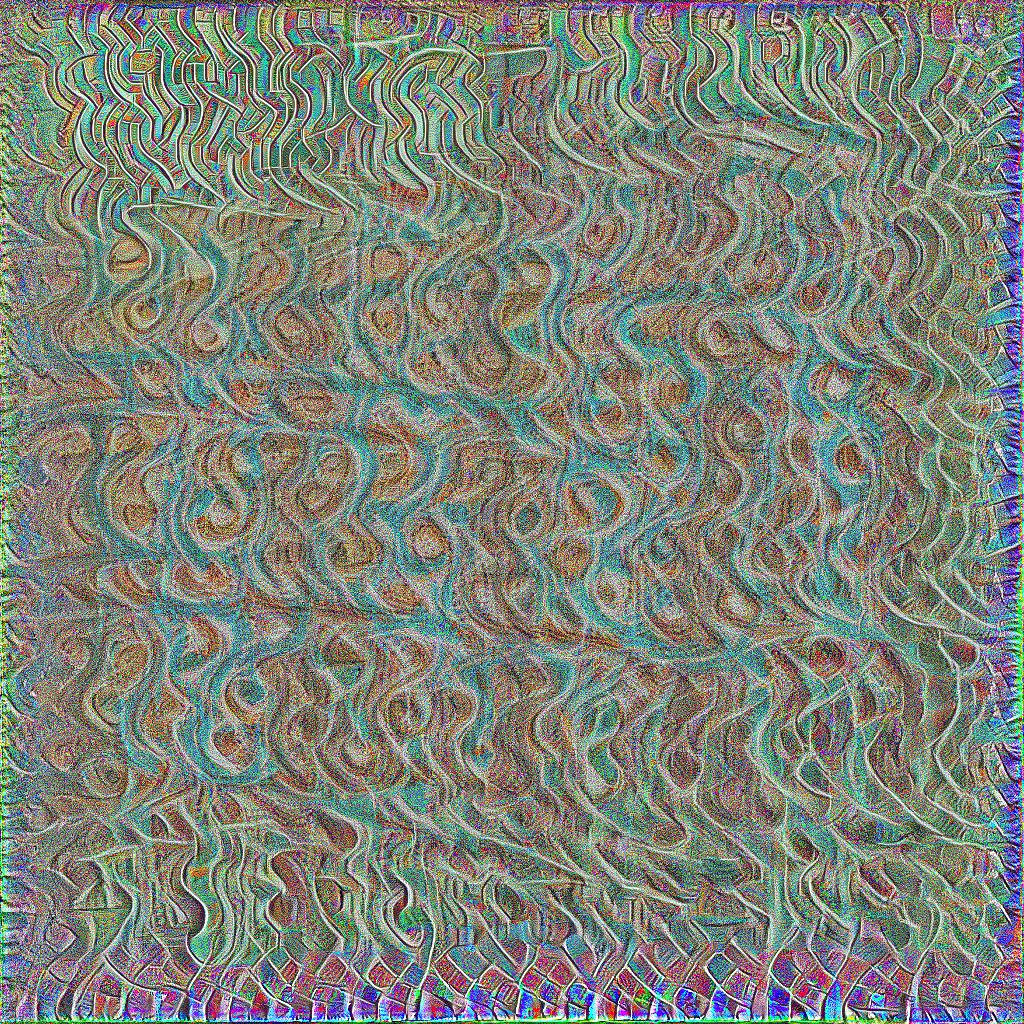}
		\caption{\texttt{b16\_conv1} and \texttt{b8\_conv1}, no normalization}
	\end{subfigure}
	\hspace*{\fill}
	\begin{subfigure}{0.3\textwidth}
		\includegraphics[width=\linewidth]{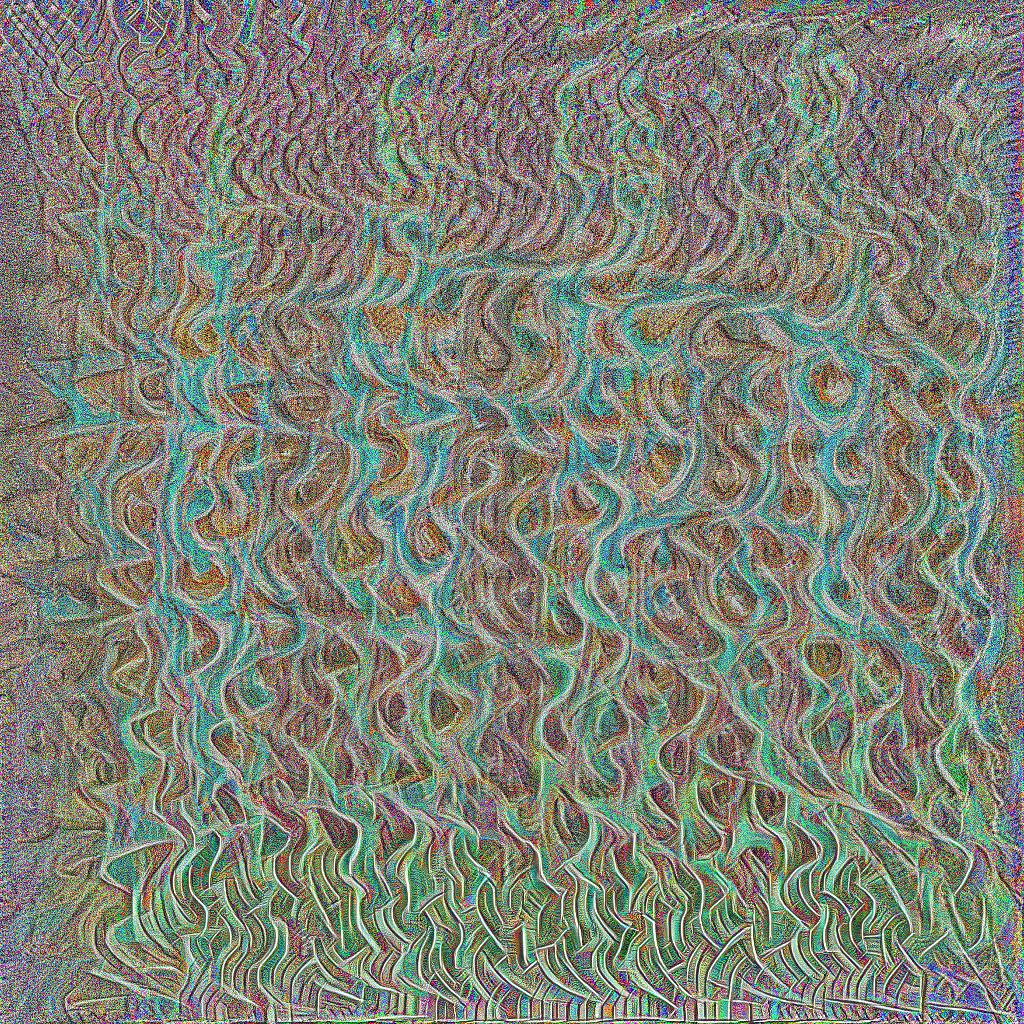}
		\caption{\texttt{b16\_conv1} and \texttt{b8\_conv1}, with normalization}
	\end{subfigure}
	\hspace*{\fill}
	\begin{subfigure}{0.3\textwidth}
		\includegraphics[width=\linewidth]{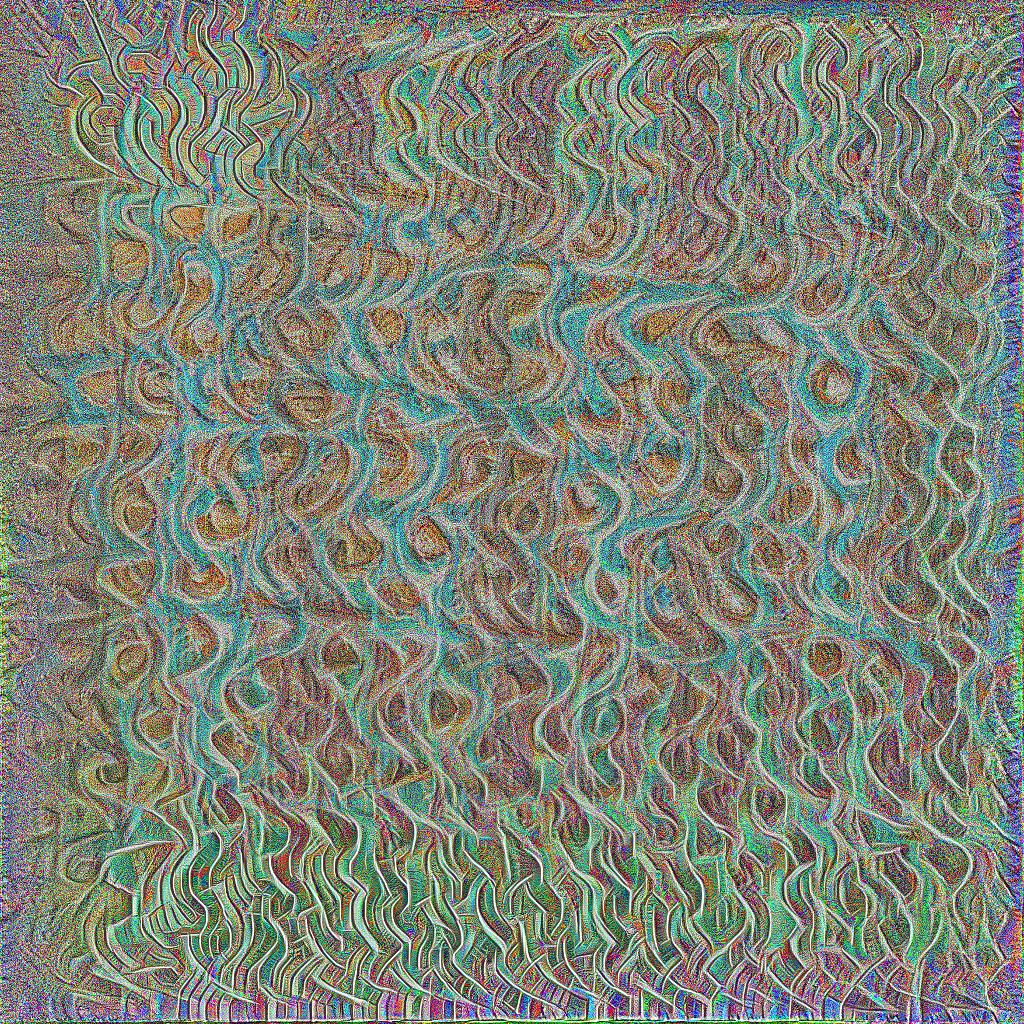}
		\caption{\texttt{b16\_conv1} and \texttt{b8\_conv1}, with square-root normalization}
	\end{subfigure}

	\caption{Results of Discriminator Dreaming starting from a random noise image, using the Earthview model and two different layers. We use $10$ octaves (and resize them), with an octave scale of $1.4$, learning rate of $0.01$, for a total of $20$ iterations. When using multiple layers, we can normalize each by either its number of elements or square-root of the number of elements, so as to let each layer have equal effect on the final image. }
	\label{fig:random_image}
\end{figure}

In lack of a starting image, as per the original DeepDream results, we can start from a random noise image (initialized with a seed), to see what the network generates. Figure \ref{fig:random_image} shows the result of Discriminator Dreaming with two different layers, starting from a random image with seed $0$.

\paragraph{Using more than one layer}

The user may also use more than one layer to play with. One of the layers may overtake the other, so we can normalize each by either the number or square root of elements of elements per layer. There is no limit on the number of layers to use, though it is recommended to start with a lower amount for better control of the final result.

\subsection{Video}\label{video}
Applying the algorithm iteratively on the outputs, as per the original DeepDream, we can create some interesting video outputs. In each case, the user can set the frames per second (FPS) and length of the video, learning rate (so as to better control the effect per frame), the number of octaves, the octave scale, which layers to use from the Discriminator, should each layer be normalized, etc. Each frame will be saved in the resulting directory, allowing for the code to save as a video in the end or for the users to use each individual frame as they please. The full code and list of available settings can be found at \url{https://github.com/PDillis/stylegan3-fun}.

We showcase some examples of zoom, rotation, and translation. For each, if the image is zoomed or rotated, the background will be black by default, but this can be changed by the user. Note that each of the following can also be $0$, so the generated video will be iteratively applying the effect to a static image.

\paragraph{Zoom} We zoom the dreamed image after each iteration by a fixed amount of pixels and then we resize it back again to the Discriminator's expected image size, forming a loop. A sample video of zooming in (by one pixel) can be found at \url{https://youtu.be/HAO3Kj2Y2Ok}. In it, we use the layer \texttt{b16\_conv1} of the MetFaces model, starting from image \texttt{00012.png} of FFHQ. We use 5 octaves whilst resizing them, a learning rate of $5\times10^{-3}$, for a total of $30$ seconds at $30$ FPS, with $10$ iterations per frame. Note that the number of pixels to zoom by can be of arbitrary magnitude and sign, so a zoom-out result can also be obtained if it is set to e.g. $-1$. 

\paragraph{Rotation} We rotate each dreamed image by a fixed angle counter\-clockwise, usually low, before feeding it back to the Discriminator (without expanding the image). A sample video of rotation (rotating by $0.2\deg$ counter-clockwise) can be found at \url{https://youtu.be/0QD1O_g-VcE}. In it, we use layer \texttt{b8\_conv1} of the MinecraftGAN model, using $5$ octaves without resizing, a learning rate of $5\times10^{-3}$, for a total of $60$ seconds at $30$ FPS, with $20$ iterations per frame. 

The rotation angle can be higher (or negative), it can be set so that the user gets the desired result. Another sample video using both rotation (by $0.1\deg$ counter-clockwise) and zoom (by $-1$ pixel) can be found at \url{https://youtu.be/hEJKWL2VQTE}. We use the same settings as the previous video, but an FPS of $25$ as well as reversing the final video, resulting in a fake zoom-in.

\paragraph{Translation} We can translate the image in both horizontal and vertical axes in an independent manner, giving a panning result. The image is translated from left to right and from top to bottom when using positive values of translation. As an example, a sample video using both horizontal and vertical translation can be found at \url{https://youtu.be/_WCyhrymo-0}. We use the \texttt{b16\_conv1} layer of the MinecraftGAN model and translate both axes by $1$ pixel per frame, a learning rate of $5\times10^{-3}$ with $20$ iterations, using 5 octaves without resizing.
\end{document}